# Visualization of Decision Trees based on General Line Coordinates to Support Explainable Models

Alex Worland
Department of Computer Science
Central Washington University
Ellensburg, USA
Alex.Worland@cwu.edu

Sridevi Wagle
Department of Computer Science
Central Washington University
Ellensburg, USA
Sridevi.Wagle@cwu.edu

Boris Kovalerchuk
Department of Computer Science
Central Washington University
Ellensburg, USA
Boris.Kovalerchuk@cwu.edu

*Abstract*—Visualization of Machine Learning (ML) models is an important part of the ML process to enhance the interpretability and prediction accuracy of the ML models. This paper proposes a new method SPC-DT to visualize the Decision Tree (DT) as interpretable models. These methods use a version of General Line Coordinates called Shifted Paired Coordinates (SPC). In SPC, each n-D point is visualized in a set of shifted pairs of 2-D Cartesian coordinates as a directed graph. The new method expands and complements the capabilities of existing methods, to visualize DT models. It shows: (1) relations between attributes, (2) individual cases relative to the DT structure, (3) data flow in the DT, (4) how tight each split is to thresholds in the DT nodes, and (5) the density of cases in parts of the n-D space. This information is important for domain experts for evaluating and improving the DT models, including avoiding overgeneralization and overfitting of models, along with their performance. The benefits of the methods are demonstrated in the case studies, using three real datasets.

## I. INTRODUCTION

Evaluation and enhancement of Machine Learning (ML) models, including their interpretability and prediction accuracy, requires advanced tools. Visualization is an important approach for this [4]. This paper deals with visualization of trained Decision Trees (DTs) as interpretable models offering a new visualization method SPC-DT.

The motivation for developing this method is the following. The existing methods for visualizing decision trees as shown in Fig. 1 show only a general structure of the DT. It does not show how the cases of different classes are split by the DT. It requires additional information to be put next to each node: how many cases of each class are going to this node and/or to put a small distribution diagram as shown in Fig. 2.

The representation in Fig. 1 allows for the tracing of a classification prediction from the root to the leaf for a new n-D point to be predicted. However, this is not sufficient to describe how good this prediction is. The answer needs to be derived from additional knowledge like the purity of each node, DT accuracy, and so on. Fig. 2 from [2] shows a part of this information.

It is well known that the **thresholds** at nodes of the DT are **not unique**. Shifting thresholds within some limits can keep the DT prediction on training and/or validation data the same. However, a new borderline case with values near the thresholds can be vulnerable due to this variability of the thresholds. The prediction can dramatically change class, making it **unstable**. The existing decision tree visualization methods are weak in the analysis of such situations and predictions can be vulnerable to being neither stable nor meaningful.

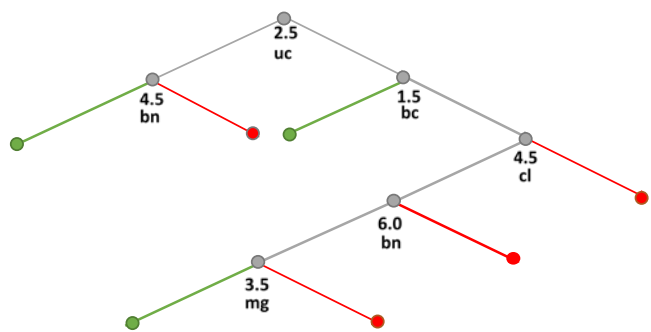

Fig. 1. Traditional visualization of Wisconsin Breast Cancer (WBC) data in the decision tree. Green edges and nodes indicate the benign class and red edges and nodes indicate the malignant class. Gray edges and nodes indicate cases that have not been decided at the current DT level.

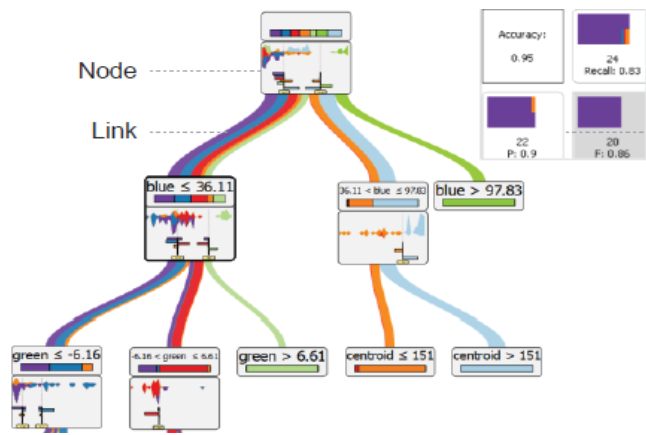

Fig. 2. DT visualization as streams [2].

A related problem is **tight thresholds**, where cases of opposite classes are close to the threshold and overlap. It is difficult to set up the right thresholds automatically. A slight change of the threshold can change classification for many training, validation, and new cases making the prediction unreliable like we already discussed above regarding new borderline cases. Thus, for the tight borders, many training and validation cases are borderline cases in contrast with the wide borders between classes.

Adjusting thresholds via interaction with visuals is a significant advantage, especially with the ability to use the domain knowledge provided by the end-users. The existing visualizations of DTs are weak in this.

The next well-known problem with DTs is **overfitting** which requires pruning DTs by shortening some branches. It cuts out the number of thresholds but makes a DT less accurate. It is difficult to balance the size of the tree and its accuracy. The visualization can help in setting up the right balance by incorporating the domain knowledge of the end-users.

Some known visualizations accompany decision trees with frequency distributions of attributes at each node, which are shown in Fig. 2, but this is difficult to scale for larger DTs.

Next, the existing visualizations do not support showing all training data with the decision tree and similarly do not show all validation data with the decision tree to compare how they are represented in the DT in detail. For instance, Fig. 2 does not allow for a clear perception of borderline cases. These drawbacks limit the abilities for deep analysis and improvement of the decision tree.

The proposed approach allows visualizing, separately or together, all training, validation, and test data in the DT. It is important to note that this visualization is lossless [9], thus, it will visualize how validation and test data are related to the training data and judge the stability and reliability of the decision tree. It also allows visualizing outlier cases.

For the decision tree model, which is already constructed, the problem is to convince the user that the DT prediction should be **trusted**. Unfortunately, the existing visualization methods are very limited in conveying confidence to the user. To address this issue, the proposed approach allows **tracing** any new case relative to the training and validation data in the decision tree. This also helps to justify or adjust the thresholds of the DT for borderline cases.

While overfitting is a common problem for decision trees, **overgeneralization** is less noticed but is still very important [10]. Consider an example where attribute $x_1$ is in the range [0,10] and a branch of the DT states that if $x_1 \leq 2.5$, then the case belongs to class 1. Also, no case of class 1 can have $x_i$ less than 1.5 and all training cases are within that range [1.5, 2.5]. The decision three overgeneralized cases of class 1 to the interval [0, 2.5]. The proposed visualization method allows for the perception of this overgeneralization and the ability to correct the range to its actual limits.

## II. SHIFTED PAIRED COORDINATES – DECISION TREE METHOD

### A. Concept and Algorithm

The concept of General Line Coordinates (GLC) [3,9] is behind the proposed method. Specifically, the method employs the Shifted Paired Coordinates (SPC) visualization technique. As the name indicates, SPC is a set of shifted pairs of Cartesian coordinates. In SPC, a directed graph (digraph) visualizes an n-D point **x** in 2-D, with nodes formed by consecutive pairs $(x_i, x_{i+1})$ of values of coordinates of **x**, connected by directed edges.

The SPC-DT method expands and complements capabilities available in the traditional visualization of DT models by visualizing: (1) *relations* between attributes, (2) *individual cases* relative to the DT, (3) *detailed data flow* in the DT, (4) the *tightness* of each split threshold in the DT nodes, and the (5) *density* of cases in parts of the n-D space. This information is of significant value to domain experts in evaluating and improving DT models. To the best of our knowledge, none of the existing DT visualization methods support all these capabilities.

The idea of the Shifted Paired Coordinates – Decision Tree (**SPC-DT**) method is presented below first conceptually, then in examples. The major steps of the SPC-DT process to generate a DT visualization from a given decision tree model are:

- Parsing the DT model.
- Pairing attributes.
- Building a set of paired Cartesian coordinates.
- Drawing each pair of coordinates in the default location shifted one over another.
- Mapping a part of the decision tree to a respective pair of coordinates as rectangles based on the threshold values.
- Color rectangles according to classes.
- Drawing selected n-D points as directed graphs in the SPC.
- Interactive modification of visualization. The options include: (1) change a set of visualized n-D points, (2) change the location of a coordinate pair, (3) change colors of the classes, (4) negate/flip coordinates, (4) swap vertical and horizontal coordinates, (5) condense points in a rectangle, (6) show overall DT structure (7) adjust thresholds, and (8) compute accuracy after adjusting thresholds.

The concept of **Shifted Paired Coordinates** (**SPC**) is explained below using an example. Consider a 6-D point $\mathbf{x} = (x_1, x_2, \ldots, x_6) = (1, 2, 3, 2, 5, 1)$. In SPC visualization, the first pair $(x_1, x_2) = (1, 2)$ is visualized as a point in the first pair of coordinates $(X_1, X_2)$. The second pair $(x_3, x_4)=(3, 2)$ is visualized in the second pair of coordinates $(X_3, X_4)$ and the last pair $(x_5, x_6)=(5, 1)$ is visualized in the third pair of coordinates $(X_5, X_6)$.

These pairs of coordinates are shifted, relative to each other to avoid overlap, then these three points are connected forming a directed graph $(1, 2) \rightarrow (3, 2) \rightarrow (5, 1)$ as seen in Fig. 3. In this way, shifted pairs of coordinates visualize any n-D point in 2-D, without loss of information [3,9].

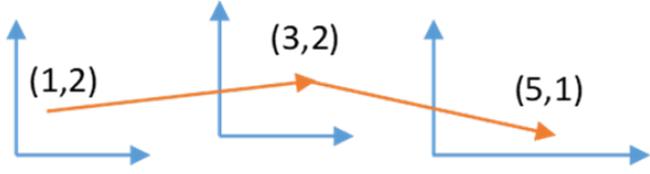

Fig.3. 6-D point x = ($x_1,x_2,…,x_6$) = (1,2,3,2,5,1) in Shifted Paired Coordinates (SPC) visualization.

*B. Example*

Below, the SPC-DT method is explained using an ID3 Decision Tree trained on 349 cases of 2 classes for Wisconsin Breast Cancer (WBC) data [1]. These cases are about 50% of all WBC Data. Fig. 4 presents this DT, Table I shows its performance, and Fig. 5 shows the traditional visualization of this DT.

```
ucellsize < 2.5
    bnuclei < 4.5 then class = benign (100.00 % of 200 cases)
    bnuclei ≥ 4.5 then class = malignant (66.67 % of 6 cases)
ucellsize ≥ 2.5
    bchromatin < 1.5 then class = benign (87.50 % of 8 cases)
    bchromatin ≥ 1.5
        clump < 4.5
            bnuclei < 6.0
                mgadhesion < 3.5 then class = benign  (100.00 % of 5 cases)
                mgadhesion ≥ 3.5 then class = malignant (66.67 % of 6 cases)
            bnuclei ≥ 6.0 then class = malignant (100.00 % of 8 cases)
        clump ≥ 4.5 then class = malignant (93.97 % of 116 cases)
```
Fig. 4. ID3 Decision Tree for WBC data

TABLE I. ID3 DT PERFORMANCE OF TREE IN FIG. 4.

| Error rate | 0.0716 | | | |
|---|---|---|---|---|
| Values prediction | Confusion matrix | | | |
| Value | Recall | 1-Precision | | begnin | malignant | Sum |
| begnin | 0.9058 | 0.0194 | begnin | 202 | 21 | 223 |
| malignant | 0.9683 | 0.1469 | malignant | 4 | 122 | 126 |
| | | | Sum | 206 | 143 | 349 |

While the WBC data contain nine attributes, this decision tree uses only five attributes, where the attribute bnuclei is used twice with two different splits. Thus, we have 5-D data in this DT that we represent as a 6-D point, with the attribute bnuclei repeated. This also helps to get the three pairs of coordinates required for SPC and SPC-DT.

The root of this tree is based on the attribute ucellsize, while the next node is based on the attribute bnuclei. For the SPC-DT visualization, these two attributes form the first pair of Cartesian coordinates, with ucellsize as a horizontal coordinate and bnuclei as a vertical coordinate. Fig. 5 shows this pair on the left. The next nodes of the DT, based on bchromatin and clump attributes, form the second pair of coordinates, as seen in the middle of Fig. 5. The last nodes, based on bnuclei and mgadhesion, form the last pair of Cartesian coordinates.

If the values of the first two coordinates (ucellsize, bnuclei) of a 6-D point are in the gray area, then it means that the decision tree did not define the class (label) of the 5-D case **x** yet when the case is traced in the DT. Similarly, a gray area in the second pair of coordinates is the area where the decision tree did not define the class yet and must be assessed at the next DT level. The green areas in each pair of coordinates are areas where the DT has determined that the class is "benign". Similarly, the red areas are the area, where the DT has determined that the class is "malignant". For example, the first two nodes of DT represent the rule:

If ucellsize < 2.5 & bnuclei < 4.5 then class = **benign**.

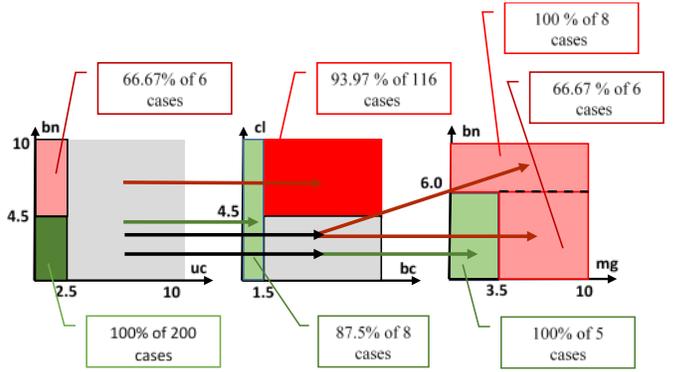

Fig. 5. WBC decision tree in Shifted Paired Coordinates. Lighter green and red colors show a lower density of cases in the respective areas.

The green area in the first pair of coordinates visualizes this rule. Similarly, the red area in this pair of coordinates represents the second rule for this DT:

If ucellsize < 2.5 & bnuclei ≥ 4.5 then class = **malignant**.

The gray area in this pair of coordinates represents cases that the DT did not classify yet. The visualization process is continued for the remaining nodes of the DT with the rules:

If ucellsize ≥ 2.5 & bchromatin < 1.5 then class = **benign** (see the green arrow from the gray area in the first plot to the green area in the middle plot).

If bchromatin ≥ 1.5 & clump ≥ 4.5 then class = **malignant** (see the red arrow from the gray area in the first plot to the red area in the middle plot).

If bchromatin ≥ 1.5 & clump < 4.5 & bnuclei < 6.0 & mgadhesion < 3.5 then class = **benign** (see the green arrow from the gray area in the middle plot to the green area in the right plot).

If bchromatin ≥ 1.5 & clump < 4.5 & bnuclei < 6.0 & mgadhesion ≥ 3.5 then class = **malignant** (see the red arrow from the gray area in the middle plot to the red area in the right plot).

If bchromatin ≥ 1.5 & clump < 4.5 & bnuclei ≥ 6.0 then class=**malignant** (see the red arrow from the gray area in the middle plot to the upper red area in the right plot).

The advantages of SCP-DT visualizations in Fig. 5 are in showing:
1. relations between *2 attributes* inside of each plot,
2. relations between *4 attributes* using green and red arrows between two plots,

3. relations between <u>all 6 attributes</u> using green and red arrows between three plots and more attributes for larger dimensions,
4. the <u>tree structure</u> and data flow by arrows in the DT.

Fig. 6 shows several 6-D points in SPC-DT that reach terminal nodes of the DT. Graphs of such individual cases in Fig. 6 allow observing how close these cases are to each split threshold in the DT nodes. Also, graphs for training cases allow observing the density of the cases in the parts of the n-D space. Using color intensity (lighter/darker) shows the density of cases in the parts of the space (see Fig. 5).

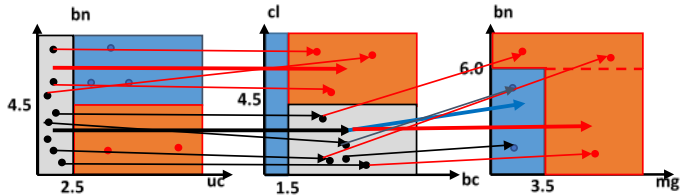

Fig. 6. WBC decision tree in SPC with several cases that reach terminal nodes of the DT.

All these features together help to increase confidence in the DT model interpretability and accuracy for future predictions. For instance, consider a DT with all actual training cases far from the threshold (border) area, but a new case to be predicted is close to this threshold. Likely, the DT model overgeneralized the training data making the predictions too risky.

In general, SPC-DT visualization supports domain experts in evaluating the DT model performance, interpretation, overgeneralization, and overfitting. The interactive version of the SPC-DT method supports turning on/off "context" attributes (attributes that are not part of the tree), but a part of the given dataset. These attributes are associated with the tree root. When such attributes are turned off, they are grayed or made semitransparent which still allows for seeing the context.

To minimize occlusion and simplify visual patterns when visualizing a large number of cases and classes, the SPC-DT method allows showing only graphs of "centers," min/max of clusters of cases with the remaining cases gray or semitransparent. Additionally, to help the user make confident predictions, the SPC-DT method allows for showing the error rate for each dataset, branch of the tree, and the full confusion matrix.

III. CASE STUDIES

Below, visualization experiments with SPC-DT on datasets from UCI ML Repository [1] with more DTs are reported.

*A. Dataset 1: Full Wisconsin Breast Cancer (WBC Original)*

Fig. 7 shows the decision tree that was used for visualizing the WBC data in SPC-DT, Table II shows the tree's performance, and Fig. 8 shows all WBC data in SPC-DT in default locations for each coordinate pair. The gray areas indicate locations within each paired coordinate space where a case's class cannot be determined at the current DT level and must be followed to the next DT level to be assessed further.

Different shades of gray are used to show differing destinations of cases in the gray areas. Each case within the same shade of gray goes to a single destination plot. The red areas indicate cases that are determined by the DT to be "malignant" while the green areas indicate cases that are determined by the DT to be "benign." These cases terminate at the current level since a class has been determined.

- ucellsize < 2.5000
  - bnuclei < 4.5000
    - clump < 6.5000 then classe = **begnin** (100.00 % of 407 examples)
    - clump >= 6.5000 then classe = **malignant** (60.00 % of 5 examples)
  - bnuclei >= 4.5000 then classe = **malignant** (52.94 % of 17 examples)
- ucellsize >= 2.5000
  - ucellsize < 4.5000
    - bnuclei < 2.5000
      - normnucl < 2.5000 then classe = **begnin** (100.00 % of 19 examples)
      - normnucl >= 2.5000 then classe = **begnin** (54.55 % of 11 examples)
    - bnuclei >= 2.5000
      - clump < 6.5000
        - bchromatin < 3.5000 then classe = **begnin** (63.64 % of 11 examples)
        - bchromatin >= 3.5000 then classe = **malignant** (83.33 % of 18 examples)
      - clump >= 6.5000
        - normnucl < 7.0000 then classe = **malignant** (100.00 % of 24 examples)
        - normnucl >= 7.0000 then classe = **malignant** (88.89 % of 9 examples)
  - ucellsize >= 4.5000
    - clump < 6.5000
      - bnuclei < 8.5000
        - bchromatin < 4.5000 then classe = **begnin** (50.00 % of 8 examples)
        - bchromatin >= 4.5000
          - mgadhesion < 8.5000 then classe = **malignant** (100.00 % of 16 examples)
          - mgadhesion >= 8.5000 then classe = **malignant** (75.00 % of 4 examples)
      - bnuclei >= 8.5000 then classe = **malignant** (100.00 % of 44 examples)
    - clump >= 6.5000 then classe = **malignant** (100.00 % of 106 examples)

Fig. 7. ID3 Decision tree for WBC data.

TABLE II. PERFORMANCE OF TREE IN FIG. 7.

| Error rate | | | 0.0401 | | |
|---|---|---|---|---|---|
| Values prediction | | | Confusion matrix | | |
| Value | Recall | 1-Precision | | begnin | malignant | Sum |
| begnin | 0.9672 | 0.0285 | begnin | 443 | 15 | 458 |
| malignant | 0.9461 | 0.0617 | malignant | 13 | 228 | 241 |
| | | | Sum | 456 | 243 | 699 |

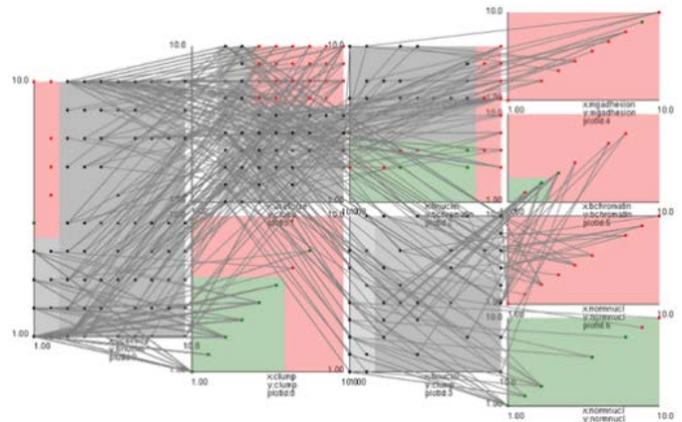

Fig. 8. WBC in default SPC-DT.

The visualization in Fig. 8 is not very informative due to the heavy overlap of lines. This makes following individual cases difficult, especially in attribute pairs with multiple gray zones. Fig. 9 and Fig. 10 show the same data with the individual coordinate pairs rearranged and relocated to make the SPC-DT visualization clearer.

Fig. 9 shows the same data as Fig. 8, but with a significant improvement in clarity due to the rearrangement and relocation of coordinate pair plots. Care has been taken to reduce the amount of line overlap, which greatly increases the clarity of the DT and the WBC cases that pass through it. However, more can be done to make the SPC-DT visualization clearer. Many lines overlap in several coordinate pair plots and many cases pass over classification areas to which they do not belong.

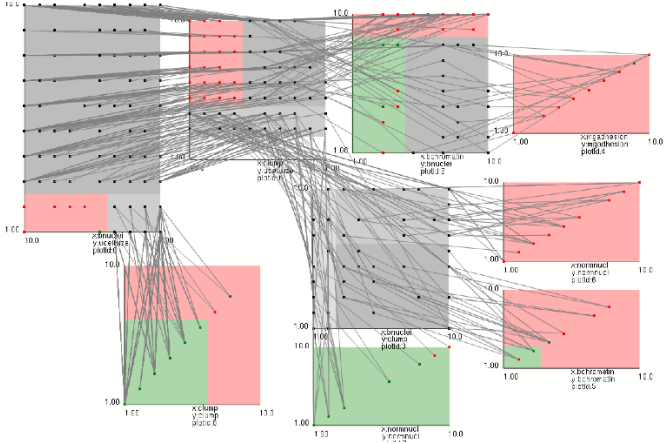

Fig. 9. WBC data in SPC-DT with rearrangement and relocation.

Fig. 10 shows the same data as Fig. 8 and Fig. 9, but with the condensation of points in gray zones, separated by each case's class. This allows for a significant increase of clarity, especially in coordinate pair plots that have a high number of cases passing through them, like the first plot of the left in Fig. 10, as well as plots that have a high number of gray zones like in other plots in Fig. 10.

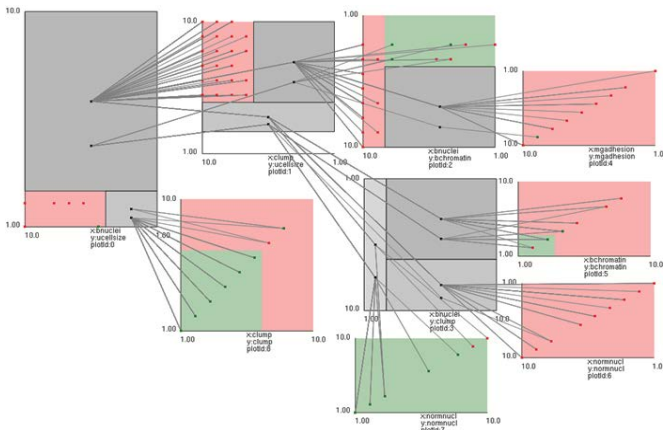

Fig. 10. WBC data in SPC-DT with rearrangement, relocation, and condensation.

However, the use of condensation naturally reduces the amount of data that can be losslessly visualized. The closer a point gets to the border of classification zones, the more uncertain the classification can be, and by condensing all points in a 2D space to a single point, this information is lost.

## B. Dataset 2: Iris Dataset

This dataset contains 150 cases of 3 classes. Each case is presented by 4 attributes. Fig. 11 shows the decision tree that was used for visualizing the Iris data in SPC-DT, Table III shows the tree's performance, and Fig. 12 shows a visualization of the DT using the SPC-DT method with all Iris data in default locations for each coordinate pair.

In Fig. 12, the gray areas indicate locations within each paired coordinate space where a case's class cannot be determined at the current DT level and must be followed to the next DT level to be assessed further. The red area indicates cases that are determined by the DT to be "Setosa," the blue area indicates cases that are determined by the DT to be "Versicolor," and the green areas indicate cases that are determined by the DT to be "Virginica." These cases terminate at the current level since a class has already been determined. Like Fig. 8, Fig. 12 does not show patterns clearly due to the heavy overlap of lines. Following individual cases is easier than in Fig. 9, but still suffers from line congestion.

- petal-length < 2.4500 then class = **Iris-setosa** (100.00 % of 50 examples)
- petal-length >= 2.4500
  - petal-width < 1.7500
    - petal-length < 4.9500
      - sepal-width < 2.5500
        - sepal-width < 2.4500 then class = **Iris-versicolor** (100.00 % of 9 examples)
        - sepal-width >= 2.4500 then class = **Iris-versicolor** (80.00 % of 5 examples)
      - sepal-width >= 2.5500 then class = **Iris-versicolor** (100.00 % of 34 examples)
    - petal-length >= 4.9500 then class = **Iris-virginica** (66.67 % of 6 examples)
  - petal-width >= 1.7500
    - sepal-length < 5.9500 then class = **Iris-virginica** (85.71 % of 7 examples)
    - sepal-length >= 5.9500 then class = **Iris-virginica** (100.00 % of 39 examples)

Fig. 11. ID3 Decision Tree for Iris data.

TABLE III. PERFORMANCE OF TREE IN FIG. 11

| Error rate | | 0.0267 | | | |
|---|---|---|---|---|---|
| Values prediction | | Confusion matrix | | | |
| Value | Recall | 1-Precision | | | |
| | | | Iris-setosa | Iris-versicolor | Iris-virginica | Sum |
| Iris-setosa | 1.0000 | 0.0000 | | | | |
| Iris-versicolor | 0.9400 | 0.0208 | Iris-setosa | 50 | 0 | 0 | 50 |
| Iris-virginica | 0.9800 | 0.0577 | Iris-versicolor | 0 | 47 | 3 | 50 |
| | | | Iris-virginica | 0 | 1 | 49 | 50 |
| | | | Sum | 50 | 48 | 52 | 150 |

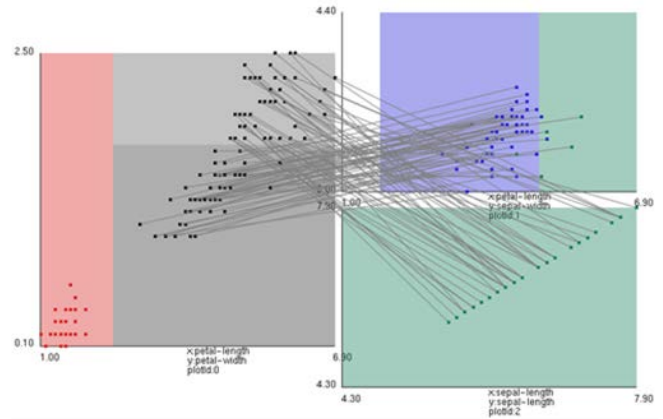

Fig. 12. Iris in default SPC-DT.

Fig. 13 shows an improvement in clarity after a simple relocation of coordinate pair plots. Relocating them in this way reduces the amount of line crossing and eliminates the paths that cross over classification zones that do not match the destination class.

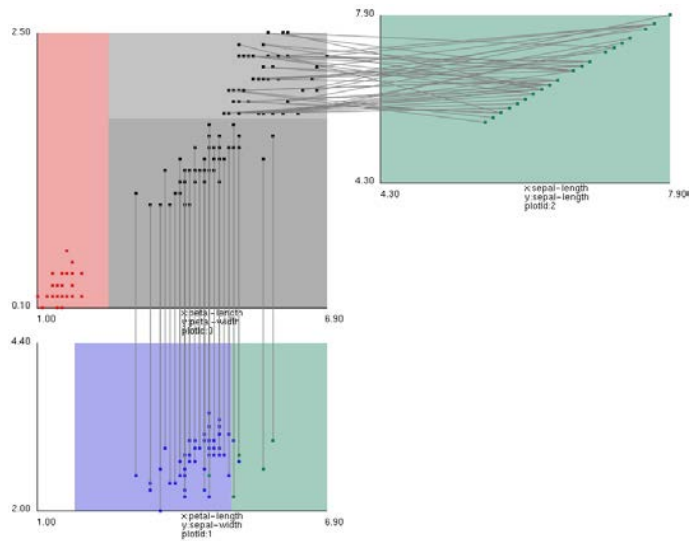

Fig. 13. Iris data in SPC-DT relocated.

Fig. 14 shows the use of condensation with the Iris data. The points outlined in red seen in plots 2 and 1 indicate cases that have been misclassified by the DT. The plots have been relocated and rearranged to take advantage of the condensation of points in gray areas of the plot on the left, which allows for better clarity of incorrectly classified cases and the path they take to their destination class in the decision tree.

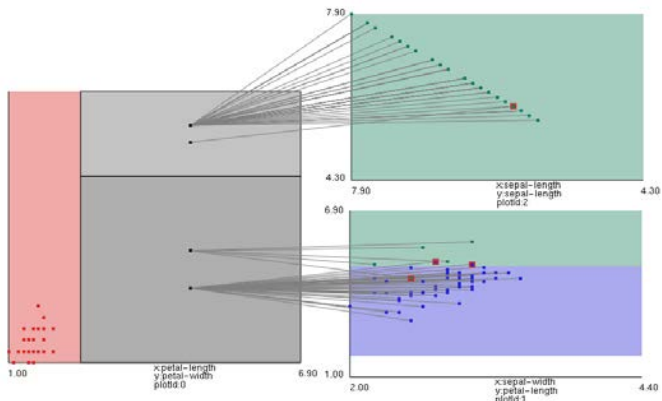

Fig. 14. Iris in SPC-DT relocated, rearranged, and condensed.

*C. Dataset 3: Wine Dataset*

This dataset contains 178 cases of 3 classes. Each case is presented by 13 attributes. Fig. 15 and Table IV describe the DT used.

- Flavanoids < 1.5750
  - Color-intensity < 3.8250 then class = **class_2** (100.00 % of 13 examples)
  - Color-intensity >= 3.8250
    - Toal-Phenols < 2.0100 then class = **class_3** (100.00 % of 42 examples)
    - Toal-Phenols >= 2.0100 then class = **class_3** (85.71 % of 7 examples)
- Flavanoids >= 1.5750
  - Proline < 724.5000
    - Malic-Acid < 3.9250 then class = **class_2** (100.00 % of 49 examples)
    - Malic-Acid >= 3.9250 then class = **class_2** (80.00 % of 5 examples)
  - Proline >= 724.5000
    - Alcohol < 13.0200 then class = **class_2** (66.67 % of 6 examples)
    - Alcohol >= 13.0200 then class = **class_1** (100.00 % of 56 examples)

Fig. 15. ID3 Decision tree for Wine data.

TABLE IV. PERFORMANCE OF TREE IN FIG. 15.

| Error rate | | | 0.0225 | | | |
|---|---|---|---|---|---|---|
| Values prediction | | | Confusion matrix | | | |
| Value | Recall | 1-Precision | | class_1 | class_2 | class_3 | Sum |
| class_1 | 0.9492 | 0.0000 | class_1 | 56 | 3 | 0 | 59 |
| class_2 | 0.9859 | 0.0411 | class_2 | 0 | 70 | 1 | 71 |
| class_3 | 1.0000 | 0.0204 | class_3 | 0 | 0 | 48 | 48 |
| | | | Sum | 56 | 73 | 49 | 178 |

Fig. 16 shows all Wine data in SPC-DT in default locations for each coordinate pair. The gray area indicates locations within each paired coordinate space where a case's class cannot be determined at the current DT level and must be followed to the next DT level to be assessed further.

The orange area indicates cases that are determined by the DT to be "class_1," the green areas indicate cases that are determined to be "class_2," and the blue area indicates cases that are determined to be "class_3." These cases terminate at the current level since a class has already been determined. Like the other case studies, the default visualization of the Wine dataset features a high degree of line overlap between coordinate pair plots. This overlap can be improved by following a similar procedure to what was used on the Iris and WBC datasets by making use of a combination of relocation, rearrangement, and condensation.

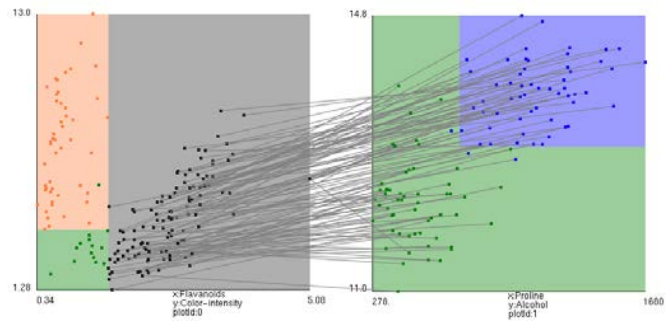

Fig. 16. Default Wine dataset in SPC-DT.

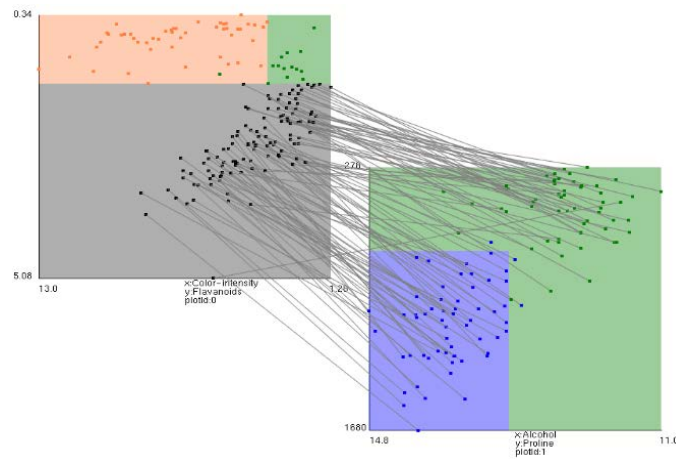

Fig. 17. Wine dataset in SPC-DT after relocation and rearrangement.

Fig. 17 shows a slight improvement in clarity after relocating the plots and rearranging their axes. This significantly reduces the number of lines passing through classification areas that do not match their destination class but does not help much to reduce the number of overlapping lines.

Fig. 18 shows the Wine data in SPC-DT after applying condensation to the gray area in the plot labeled "plotId: 0," there is a significant increase in clarity due to the reduction of line overlap. Additional clarity is gained by highlighting misclassifications, seen in Fig. 18 as points with the red frames. This can aid in understanding the performance of the decision tree being visualized.

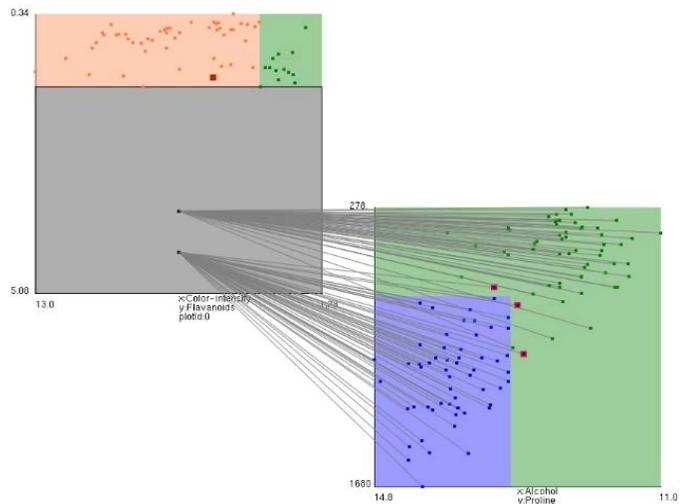

Fig. 18. Wine dataset in SPC-DT with relocation, rearrangement, and condensation.

*D. WBC Dataset with Split to Training and Validation Data*

This section illustrates the advantages of the SPC-DT method in visualizing datasets that are split into training and validation data. The split allows the observation and analysis of the relations between training and validation data that do not belong to training data.

Splitting data can increase the confidence in the DT model or otherwise can lead to updating and redesigning the decision tree, which makes the DT modeling process more efficient. Fig. 19 and Table V describe the decision tree used for visualization.

- ucellsize < 2.5000
  - sepics < 2.5000 then class = begnin (99.43 % of 349 examples)
  - sepics >= 2.5000 then class = begnin (77.42 % of 31 examples)
- ucellsize >= 2.5000
  - ucellsize < 4.5000
    - bnuclei < 3.5000 then class = begnin (78.38 % of 37 examples)
    - bnuclei >= 3.5000 then class = malignant (85.11 % of 47 examples)
  - ucellsize >= 4.5000
    - clump < 6.5000
      - bnuclei < 8.5000 then class = malignant (80.00 % of 25 examples)
      - bnuclei >= 8.5000 then class = malignant (100.00 % of 41 examples)
    - clump >= 6.5000 then class = malignant (100.00 % of 99 examples)

Fig. 19. ID3 Decision Tree of 90/10 split WBC Data.

TABLE V.   PERFORMANCE OF TREE IN FIG. 19

| Error rate | | | 0.0461 | | |
|---|---|---|---|---|---|
| Values prediction | | | Confusion matrix | | |
| Value | Recall | 1-Precision | | begnin | malignant | Sum |
| begnin | 0.9709 | 0.0408 | begnin | 400 | 12 | 412 |
| malignant | 0.9217 | 0.0566 | malignant | 17 | 200 | 217 |
| | | | Sum | 417 | 212 | 629 |

Fig. 20 and Fig. 21 show visualizations of validation and training data respectively. In these visualizations, darker shades of classification colors are used to show the number of cases that the DT rule covers. Darker colors indicate more cases. The white borders around certain cases are to improve visibility. They do not show additional information about the DT. To aid in the visualization of misclassified cases, overlapping classifications have been spread out. Tightly packed cases that appear in an upper left to lower right diagonal line all contain the same value, but they have been spread to better visualize misclassified cases that may be hard to see otherwise. A confusion matrix has been included in Fig. 20 because the confusion matrix in Fig. 19 is for the 90% training dataset.

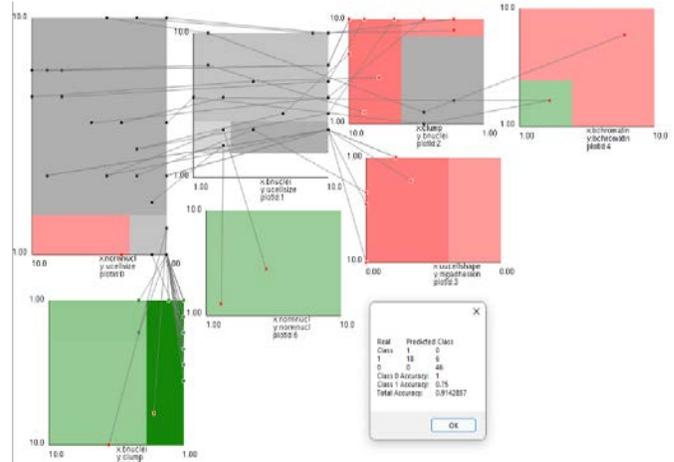

Fig. 20. Testing data in 90/10 split DT in SPC-DT.

Notably, the structure of the decision tree differs from that of Fig. 8 through Fig. 10. This is due to the DT in Fig. 20 and Fig. 21 being built using only 90% of the WBC dataset, while the DT used in Fig. 8 through Fig. 10 is built using 100% of the WBC data. This difference demonstrates instability of the DT models. While DT instability is a well known their property it is not easy to see the diference in the tradtional DT descrptions like shown in Figs. 7 and 19 in contrast by comparing figures like Figs. 10 and 21.

It is visible that the distribution of the cases in the validation data differs from that of the training data. It does not cover significant areas, which were covered by the training data. This means that the validation data may not well represent the training data and the accuracy of the validation data can not be representative of the training data. This can be seen in the difference between Fig. 20 and Fig. 21, resulted in different accuracies: 91.43% on valdation data and 95.39% on traning data.

Fig. 21 also shows the blue case, which starts in the gray area in the left plot very close to the green area. It goes to the gray area in the second plot and then terminates in the red area in the bottom plot, again very close to the green area. This visual analysis of the case shows that this classification is not reliable. A slight change of the borderlines will change the class of this case. This demonstrates the advantages of visual analysis and the SPC-DT visualization approach.

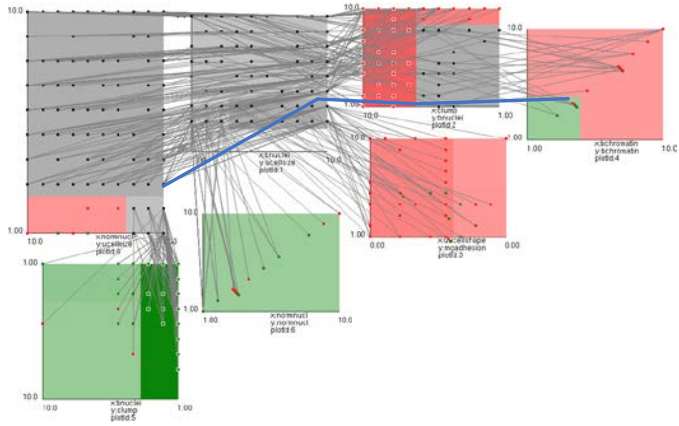

Fig. 21. Training data in 90/10 split DT in SPC-DT.

## IV. Comparison and Conclusion

About 200 different DT visualizations were reported in 2011 [8], and now Treevis.net shows over 300. These visualizations include [6,7]. While these are impressive numbers, only a few of these DT visualizations are detailed enough to represent the machine learning models. For instance, a Treemap DT visualization, which is very powerful for showing large complex trees, does not show split thresholds in the DT machine learning models. Typically, alternative DT visualizations in machine learning show the number of cases in each node and/or one-dimensional (marginal) distribution of the cases.

One of the major advantages of the SPC-DT method is its ability to show the actual cases, and their 2-D distributions, in the $(X_i, X_j)$ coordinates presented in the adjacent nodes of the DT. It allows the tracing of misclassified cases to improve the accuracy of the DT model interactively. For instance, when misclassified cases are close to the threshold of some nodes, a user can interactively move the thresholds in these nodes in the SPC-DT visualization to change the classification of those cases and see how it affects the classification of other training and validation cases.

In the SPC-DT method, a user can interactively optimize the DT. While some DT visualizations such as [5] allow the use of 1-D distributions for interactive optimization, SPC-DT allows it in more detail in 2-D distributions. In addition, SPC-DT provides a more compact visualization of a smaller number of tracing points. For instance, in Fig. 5 and Fig. 6, tracing the 6-D case (5, 8, 3, 4, 6) would require 3 tracing points while in Fig. 1 and Fig. 2 it would require 5 tracing points.

This paper proposed a new DT visualization method for machine learning. The features and advantages of this method include showing relations between multiple pairs of attributes in each plot, individual cases, the closeness of the cases to the split thresholds in the DT nodes, and the density of the cases in the parts of the n-D space. These features together help in gaining confidence in the DT model interpretability and accuracy for future predictions, avoiding overgeneralization and overfittings.

It is a significant challenge for Shifted Paired Coordinates to select an efficient pairing of coordinates that will allow for the discovery of patterns of cases of different classes clearly. Exploration of all possible pairings is not feasible, so a practical approach to avoid combinatorial exploration is applied, using a decision tree to define the pairs of shifted paired coordinates as this paper illustrates with the SPC-DT approach. The pairs are constructed using adjacent coordinates in the decision tree. In each branch of the DT, the coordinate that is at the root is paired with the coordinate in the second node of the branch. Next, the 3rd and 4th nodes form the second pair of coordinates, and so on. Some coordinates can be used in the DT several times with different thresholds. Respectively, the DT-based pairing will use them several times. This means that the decision tree approach allows us to expand the number of pairing alternatives beyond using each coordinate only in one pair. Thus, the approach proposed in this paper allows (1) efficient visualization of the decision tree, (2) its improvement using this visualization, (3) communicating the DT to end-users, and (4) finding an efficient pairing of coordinates in Shifted Paired Coordinates, which makes detecting patterns in the data much clearer. This opens an opportunity to use the SPC-DT approach beyond decision trees.